\pdfoutput=1
\documentclass[11pt]{article}

\usepackage[]{emnlp2021}

\usepackage{times}
\usepackage{latexsym}
\usepackage{graphicx}
\usepackage{multirow}
\usepackage{booktabs}

\usepackage[T1]{fontenc}
\usepackage[utf8]{inputenc}

\usepackage{microtype}

\newcommand{\waitk}{\texttt{wait-k} }
\newcommand{\K}{\texttt{k}}

\title{Simultaneous Multi-Pivot Neural Machine Translation}

\author{Raj Dabre$^1$ \quad
        Aizhan Imankulova$^2$ \quad
        Masahiro Kaneko$^3$ \quad
        Abhisek Chakrabarty$^4$ \\
        National Institute of Information and Communications Technology$^{1,4}$ \\
        CogSmart$^2$ \quad
        Tokyo Institute of Technology$^3$ \\
        {\tt $^1$raj.dabre@nict.go.jp} \quad
        {\tt $^2$aizhan.imankulova@cogsmart-global.com} \\
        {\tt $^3$masahiro.kaneko@nlp.c.titech.ac.jp} \quad {\tt $^4$abhisek.chakra@nict.go.jp} \quad
}

\begin{document}
\maketitle
\begin{abstract}
Parallel corpora are indispensable for training neural machine translation (NMT) models, and parallel corpora for most language pairs do not exist or are scarce. In such cases, pivot language NMT can be helpful where a pivot language is used such that there exist parallel corpora between the source and pivot and pivot and target languages. Naturally, the quality of pivot language translation is more inferior to what could be achieved with a direct parallel corpus of a reasonable size for that pair. In a real-time simultaneous translation setting, the quality of pivot language translation deteriorates even further given that the model has to output translations the moment a few source words become available. To solve this issue, we propose multi-pivot translation and apply it to a simultaneous translation setting involving pivot languages. Our approach involves simultaneously translating a source language into multiple pivots, which are then simultaneously translated together into the target language by leveraging multi-source NMT. Our experiments in a low-resource setting using the N-way parallel UN corpus for Arabic to English NMT via French and Spanish as pivots reveals that in a simultaneous pivot NMT setting, using two pivot languages can lead to an improvement of up to 5.8 BLEU.

\end{abstract}

\section{Introduction}
Neural machine translation (NMT) \cite{bahdanau15} is an end-to-end approach that is known to give the state of the art results for a variety of language pairs. Unfortunately, most language pairs have little to no parallel corpora, which are indispensable for NMT. In such situations, unsupervised, zero-shot, and zero-resource NMT approaches have been developed, which have enabled translation between such pairs. In particular, the most straightforward and most effective approach is the pivot language NMT approach which involves using a pivot language such that the source language sentence is translated into the pivot language, which is then translated into the target language. Pivot language translation quality is often poorer than the direct translation quality, even if a small amount of direct parallel corpus exists. 
Naturally, it is expected that pivot translation quality further degrades in a real-time translation setting where simultaneous NMT (SNMT) \cite{gu-etal-2019-improved,ma-etal-2019-stacl, Zheng_2019} approaches are used. 

SNMT approaches are those that perform a translation with partial source language sentences, which are fed to the model a word/token at a time. The most straightforward simultaneous NMT approach is the $\waitk$ approach \cite{ma-etal-2019-stacl} where the model starts generating translations, a word at a time when $K$ source words/tokens are available. Note that simultaneous NMT approaches already exhibit more inferior translation quality as compared to their non-simultaneous counterparts \cite{gu-etal-2019-improved,ma-etal-2019-stacl,zhang-etal-2020-improving}. Naturally, simultaneous pivot language NMT, where both the source to pivot and pivot to target language models are simultaneous NMT models, is expected to suffer from a drastically reduced translation quality. Fortunately, nothing prevents us from using multiple pivot languages.

\begin{figure}
    \includegraphics[scale=0.35]{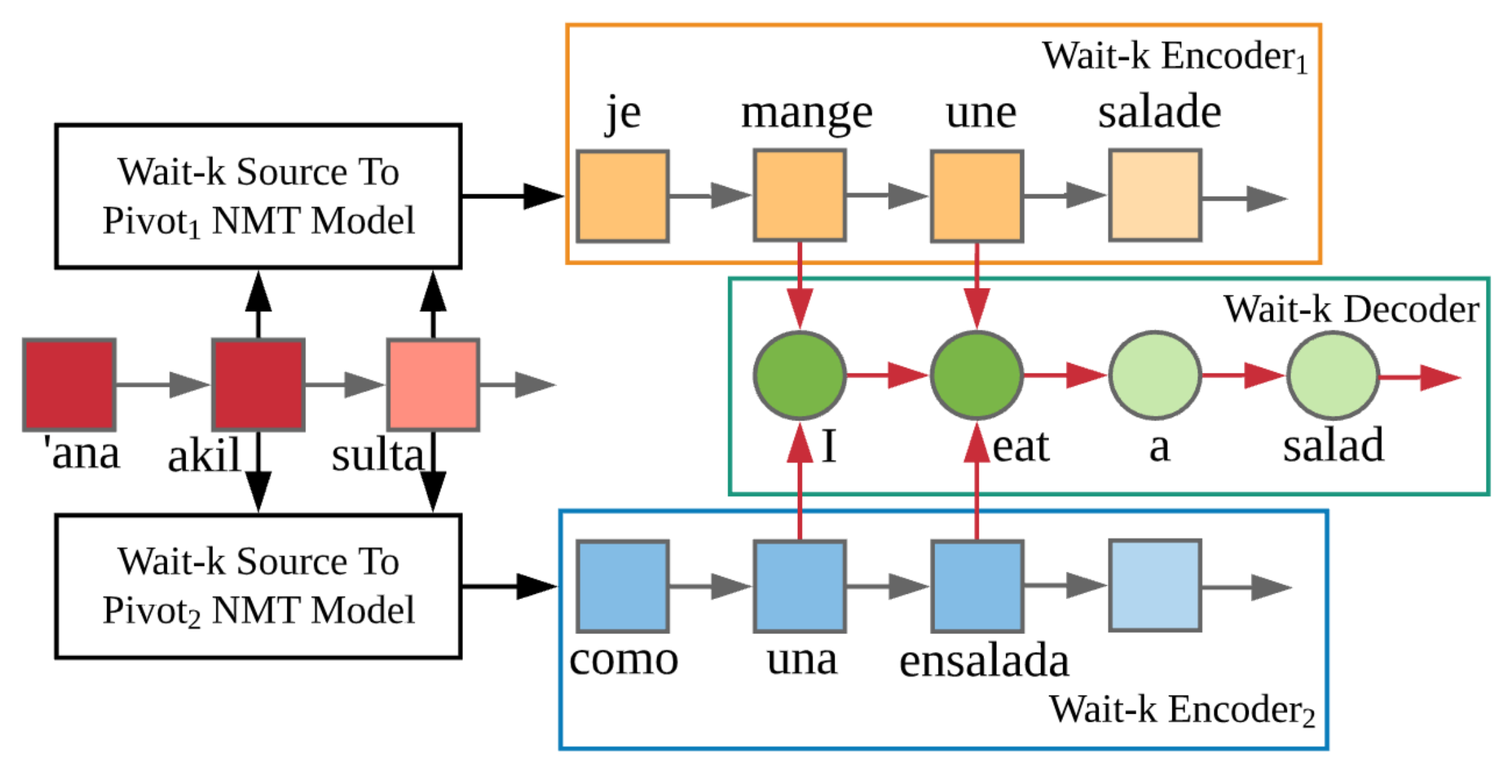}
    \caption{Our multi-pivot NMT architecture. When each component model is an SNMT model we obtain a simultaneous multi-pivot NMT (SMPNMT) model.}
    \label{fig:smpnmt}
\end{figure}

Previous work in statistical machine translation (SMT) \cite{utiyama2007, bertoldi2008phrasebased} has shown that using multiple pivot languages can significantly improve translation quality as opposed to using a single pivot language \cite{dabre2015leveraging}. In SMT, phrase-table triangulation can be done for each pivot language leading to a new phrase table following which all phrase tables are combined using an interpolation approach. In the case of NMT, however, we can resort to multi-source neural machine translation (MSNMT) \cite{Dabre-MTS2017,N16-1004} which can be helpful. MSNMT is a variation of NMT which involves translating a sentence written in two different languages into a target language. The premise of MSNMT is that each source language has phenomena that are hard to translate. Thus by providing multiple sources, the decoder is able to better disambiguate the hard-to-translate phenomena in the individual sources leading to an improvement in translation quality. In the pivot language NMT setup, if the pivot to target model is replaced with an MSNMT model, then we can achieve multi-pivot NMT (MPNMT). Finally, when the source to pivot and multi-pivot to target models are simultaneous NMT models, we arrive at simultaneous multi-pivot NMT (SMPNMT), which is the main idea in this paper. Our experiments on Arabic to English translation via French and Spanish as pivot languages show that multi-pivot NMT significantly outperforms single-pivot NMT in a non-SNMT setting and in an SNMT setting, SMPNMT outperforms single-pivot SNMT by over 5.8 BLEU. This paper is intended to be a proof of concept that multi-pivot NMT works and is useful. We leave experimentation on a variety of other language pairs in realistic settings for the future.

\section{Related Work}
The ideas in this paper revolve around pivot language NMT, SNMT and MSNMT.

Pivot language translation helps improve the accuracy of low/zero-resource translation. In the case of SMT, pivot translation was done using chaining, or phrase table triangulation \cite{wu2007pivot,utiyama2007}. \citet{dabre2015leveraging} showed that in SMT, leveraging multiple pivot languages can significantly improve the translation quality which was one of the motivations for our work. In NMT, although zero-shot translation \cite{johnson17} is an option, chaining is much simpler to use and gives reasonable results \cite{firat16b}, and hence we rely on it in this paper. \citet{cheng2017joint} first proposed the pivot based neural machine translation by jointly training source to pivot and pivot to target models, which improves the effectiveness of chaining. \citet{currey-heafield-2019-zero} proposed an alternative way of using a monolingual pivot-language data for zero-resource NMT via back-translation \cite{sennrich-haddow-birch:2016:P16-11}.

SNMT work are mainly divided into two groups based on their policies \cite{Zheng_2019}: SNMT with fixed policy \cite{dalvi-etal-2018-incremental, ma-etal-2019-stacl, Zheng_2019} and SNMT with adaptive policy \cite{cho2016can,gu2017learning,zheng2019simpler,zheng-etal-2019-speculative,zheng-etal-2020-opportunistic}.
Models with adaptive policy dynamically decide whether the model should READ or WRITE tokens with some waiting criteria. 
On the other hand, models with the fixed policy have much simpler architecture and lower latency compared to more complicated models with the adaptive policy.
As a study utilizing additional information for SNMT, it has been shown that the image information related to the translated sentence contributes to performance improvement~\cite{imankulova-etal-2020-towards,caglayan-etal-2020-simultaneous}.

\citet{N16-1004} first proposed MSNMT using multiple encoders for each source language and a single decoder for the target language. \citet{Dabre-MTS2017}, on the other hand, showed that simply concatenating multiple source sentences is sufficient, which helps avoid the need for modification of the NMT architecture. However, this prevents the special handling of individual languages and thus, we rely on the multi-encoder approach. Other multi-source NMT papers include using mixture-of-experts \cite{C16-1133}, incorporating linearized parses of
the source data \cite{currey-heafield-2018-multi}, and addressing missing data in multi-source MT \cite{nishimura-etal-2018-multi}. One major limitation of MSNMT is that N-way parallel corpora are required, and although \citet{nishimura-etal-2018-multi} partially address this issue, it is important to note that MSNMT can be achieved without any N-way corpora \citet{firat16b}. Our idea currently requires N-way corpora simply to establish its efficacy, but we will relax this constraint in the future.

\section{Background}
We briefly introduce some background concepts related to pivot NMT, SNMT and MSNMT.

\subsection{Pivot NMT}
The most straightforward approach for pivot-language NMT involves chaining two NMT models, one for source to pivot language translation and one for a pivot to target language translation. There are other sophisticated approaches such as joint training both models, but our objective is to enhance the quality of the chaining approach via multiple pivots.

\subsection{SNMT}
The most straightforward approach for SNMT is the $\waitk$ approach \cite{ma-etal-2019-stacl} with a fixed policy. As tokens are fed to the encoder one at a time, we have to rely on a unidirectional encoder that cannot attend to future tokens. Once the encoder has been fed $K$ tokens, the decoder starts generating a token at a time. This means that at a given time, the $i^{th/st}$ decoding step can only see $K+i-1$ encoder token representations. 
Once the whole input sentence is available, $\waitk$ behaves like regular NMT except with a unidirectional encoder.

\subsection{MSNMT}
A standard encoder-decoder NMT model can be converted to an MSNMT model by having $N$ encoders, one for each source language. Consider the case of 2 source languages with 2 encoders. Consider source sentences $s_{1}$ and $s_{2}$ with lengths $l_{1}$ and $l_{2}$. Suppose that the encoder representations are $E_{s_{1}}$ and $E_{s_{2}}$ with dimensions $(l_{1}, h)$ and $(l_{1}, h)$ respectively where $h$ is the hidden size. For each time step in the decoder, we first compute attentions for each source sentence independently as $A_{s_{1}}$ and $A_{s_{2}}$ with hidden sizes $h$. $A_{s_{1}}$ and $A_{s_{2}}$ are then passed through a gating mechanism to determine their relative importance and generate the final attention $A$. $A = w*A_{s_{1}} + (1-w)*A_{s_{2}}$ where $w={\rm sigmoid}(W[A_{s_{1}}:A_{s_{1}}] + b)$. $W$ is a weight matrix of dimensions $(2h,h)$ and $b$ is a bias of dimension $h$. $:$ represents concatenation. 

An MSNMT model typically requires N-way corpora for training which we use in this paper, but this need not be the case as shown by \cite{firat16b}. An MSNMT model can be converted into its $\waitk$ SNMT equivalent by enforcing the unidirectional constraint on each encoder. The decoder generates a token at a time once $K$ tokens have been fed to both encoders. In reality, each encoder can receive tokens at different rates, but for simplicity, we assume that they receive tokens at the same time.

\section{MPNMT and SMPNMT}

Figure~\ref{fig:smpnmt} gives an overview of our approach. Combining the SNMT with MSNMT lead us to multi-pivot NMT (PNMT). MPNMT involves translating the source into N pivot languages via bilingual or multilingual NMT models and then translating these N pivot languages into the target language via an MSNMT model. The bilingual or multilingual NMT models are trained using parallel corpora, and the MSNMT models are trained using N-way corpora (although this can be achieved without bilingual corpora too \cite{firat16b}). Replacing the source to pivot and pivot to target models in the MPNMT framework with their ($\waitk$) SNMT equivalents leads us to simultaneous MPNMT (SMPNMT).  

\section{Experimental Settings}
\subsection{Datasets and preprocessing}
We use the UN dataset and focus on Arabic to English translation using French and Spanish as pivot languages for our experiments. The UN corpus is N-way parallel for the training, development and test set splits. As our purpose is not to establish state-of-the-art results but to validate the efficacy of our multi-pivot method, we consider a sample of 200,000 training sentences that are also N-way parallel. This also helps simulate a low-resource setting where multi-pivot translation should be extremely beneficial. We do not perform any pre-processing on our own and let our implementation handle it via its internal mechanisms.

\subsection{Implementation, Training and Evaluation details}
For our experiments, we modify the implementation of the transformer model in the tensor2tensor toolkit\footnote{\url{https://github.com/tensorflow/tensor2tensor}} (v1.15.4). The tensor2tensor toolkit has internal tokenization and sub-word segmentation mechanisms, which we use with a subword vocabulary of size 16,000. The subword vocabularies for the encoder and decoder are separate. In the case of multiple pivot languages (multi-source models) the encoder is shared\footnote{Simply put, in the French+Spanish multi-source (pivot to target model) model, the subword vocabulary is shared between French and Spanish but the English target vocabulary is separate.} between the pivot languages which helps cut down the number of parameters. We use hyperparameter configurations corresponding to the ``transformer\_base'' model. We save and evaluate our models on the development set every 1000 batches with BLEU \cite{Papineni} as the evaluation metric. We train our models till the BLEU score does not grow for 10 consecutive evaluations. We average the last 10 saved checkpoints and then decode the model. As we are working in a simultaneous translation setting, using greedy search makes sense as tokens should be output one at a time \footnote{It's possible to consider a sophisticated beam search method, but that is beyond the scope of this paper.}.
\subsection{Models to Compare}
We train and evaluate full sentence and $\waitk$ models for $\K$=1,2,4,6 and 8 for the following models.
\begin{itemize}
    \item Direct Baselines: Direct Arabic to English.
    \item Pivot Baselines: Arabic to pivot to English using a single pivot of French or Spanish.
    \item Our proposed models: Arabic to French+Spanish to English using both French and Spanish as pivots.
\end{itemize}
All source to pivot models are unidirectional bilingual models, although it is possible to train a single one-to-many model to translate from a source to multiple pivots. When the source and pivot models are both $\waitk$, there are a total of 25 combinations of $\waitk$ for source to pivot and pivot to target. The effective $\waitk$ is the sum of the $\waitk$'s of individual models.

\section{Results}

\begin{table}[]
\centering
\scalebox{.85}{%
\begin{tabular}{cccc}
\toprule
\multicolumn{1}{l}{} & \textbf{Type} & \textbf{\begin{tabular}[c]{@{}c@{}}Pivot\\ Language\end{tabular}} & \textbf{BLEU} \\ 
\midrule
\multicolumn{1}{c}{\textbf{1}} & \textbf{Direct} & N/A & 34.0 \\ 
\multicolumn{1}{c}{\textbf{2}} & \textbf{Pivot} & French & 27.5 \\
\multicolumn{1}{c}{\textbf{3}} & \textbf{Pivot} & Spanish & 25.1 \\ 
\multicolumn{1}{c}{\textbf{4}} & \textbf{\begin{tabular}[c]{@{}c@{}}Pivot (Ours)\end{tabular}} & \begin{tabular}[c]{@{}c@{}}French+Spanish\end{tabular} & 29.9 \\ 
\bottomrule
\end{tabular}%
}
\caption{BLEU scores for baseline Arabic to English models. Results are shown for (1) direct Arabic to English, (2) Arabic to English via French as a pivot, (3) Arabic to English via Spanish as a pivot  and (4) Arabic to English via both French and Spanish as a pivot.}
\label{tab:full-sent-multipivot}
\end{table}

\subsection{Full Sentence Results}
We first show the results for full-sentence direct, single-pivot and multi-pivot NMT models in Table~\ref{tab:full-sent-multipivot}. It is clear that when compared to directly translating into the target language (row 1), translating through a pivot language (rows 2 and 3) dramatically hurts the translation quality. The translation quality drops by 6.5 and 8.9 BLEU for French and Spanish as pivots, respectively.  However, when both pivot languages are used together, the translation quality shows a significant improvement of 2.4 BLEU compared to the best translation score obtained using French as the pivot. In other words, the multi-pivot model is only 4.1 BLEU below the direct model as opposed to 6.5 to 8.9 BLEU below the direct model when single pivots were used. This result is important because the pivot language sentence is inherently noisy and translating into the target language makes it even noisier. However, combining two noisy translations via multi-source NMT can significantly enhance the translation quality. Note that the direct translation model score is the upper limit of translation quality if a parallel corpus is available. Hence, it acts as a reference upper limit score that we wish to approach via multi-pivot translation.

\subsection{Simultaneous NMT Results}

Table~\ref{tab:waitk-multipivots} contains results for simultaneous models for Arabic to English translation directly and via pivots. 

\subsubsection{SNMT vs Full Sentence Models}
First, when comparing the direct $\waitk$ results against direct full-sentence results, it can be seen that as the value of $\K$ increases, the gap between full-sentence and SNMT drops rapidly, and when $\K$ is 8, the gap drops to 2.9 BLEU. Depending on the cost-benefit requirement, we can increase or decrease the value of $\K$ to yield the desired translation quality. Next, comparing $\waitk$ pivot models to their non $\waitk$ equivalents, we can observe that the $\waitk$ equivalents give significantly more unsatisfactory results. Although this is expected, it must be noted that the drop in translation quality is even more severe (up to 6.2 BLEU (21.3 vs 27.5 with French as a pivot)) compared to the drop for direct translation models (only 2.9 BLEU). However, this is where using multiple pivots comes to the rescue. Compared to the multi-pivot full-sentence model, the $\waitk$ equivalent is only 2.8 BLEU points weaker. Thus it is clear that by employing multiple pivots, we can reduce the BLEU drop from 6.2 BLEU (using one pivot) to 2.8 BLEU (using two pivots). We suppose that using additional pivots may help close this gap even further and will verify it in the future.

\begin{table}[]
\centering
\scalebox{.85}{%
\begin{tabular}{lcrrrrr}
\toprule
\textbf{\K} & \textbf{} & \multicolumn{1}{c}{\textbf{1}} & \multicolumn{1}{c}{\textbf{2}} & \multicolumn{1}{c}{\textbf{4}} & \multicolumn{1}{c}{\textbf{6}} & \multicolumn{1}{c}{\textbf{8}} \\ \cline{2-7} 
 Direct & \textbf{} & \multicolumn{1}{c}{19.1} & \multicolumn{1}{c}{22.5} & \multicolumn{1}{c}{26.3} & \multicolumn{1}{c}{30.0} & \multicolumn{1}{c}{\textbf{31.1}} \\ 
 \midrule
 \midrule
 Pivot &  & \multicolumn{5}{c}{\textbf{P2T-\K}} \\ 
 \midrule
\multirow{6}{*}{French} & \textbf{S2P-\K} & \multicolumn{1}{c}{\textbf{1}} & \multicolumn{1}{c}{\textbf{2}} & \multicolumn{1}{c}{\textbf{4}} & \multicolumn{1}{c}{\textbf{6}} & \multicolumn{1}{c}{\textbf{8}} \\ \cline{2-7} 
 & \textbf{1} & 7.3 & 9.3 & 11.9 & 14.1 & 15.4 \\  
 & \textbf{2} & 8.4 & 10.6 & 13.7 & 16.1 & 17.3 \\  
 & \textbf{4} & 9.6 & 12.3 & 16.1 & 18.7 & 19.8 \\  
 & \textbf{6} & 9.9 & 12.8 & 16.8 & 19.6 & 21.0 \\  
 & \textbf{8} & 10.1 & 12.9 & 16.8 & 19.7 & \textbf{21.3} \\ 
 \midrule
\multirow{6}{*}{Spanish} & \textbf{1} & 5.8 & 8.2 & 11.0 & 13.0 & 14.6 \\ 
 & \textbf{2} & 6.6 & 9.3 & 12.3 & 14.7 & 16.3 \\ 
 & \textbf{4} & 7.3 & 10.4 & 13.9 & 16.4 & 18.0 \\  
 & \textbf{6} & 7.7 & 11.0 & 14.9 & 17.1 & 19.1 \\  
 & \textbf{8} & 7.6 & 11.1 & 15.1 & 17.9 & \textbf{20.0} \\
 \midrule
 & \textbf{1} & 10.4 & 12.4 & 15.1 & 16.7 & 18.3 \\  
 French+ & \textbf{2} & 11.7 & 14.1 & 17.3 & 18.7 & 20.0 \\ 
 Spanish & \textbf{4} & 13.9 & 16.6 & 20.1 & 21.6 & 23.1 \\ 
 (Ours) & \textbf{6} & 14.4 & 17.6 & 21.3 & 23.0 & 25.2 \\ 
 & \textbf{8} & 14.4 & 18.1 & 21.9 & 23.8 & \textbf{27.1} \\ 
 \bottomrule
\end{tabular}%
}
\caption{BLEU scores for $\waitk$ Arabic to English models. The values of $\K$ are 1, 2, 4, 6 and 8. In the case of pivot models, two $\waitk$ models are involved leading to 25 combinations. S2P and P2T are short for source to pivot and pivot to target. Results are shown for (1) Direct Arabic to English, (2) Arabic to English via French as a pivot, (3) Arabic to English via Spanish as a pivot and (4) Arabic to English via both French and Spanish as a pivot.}
\label{tab:waitk-multipivots}
\end{table}

\subsubsection{SNMT pivot vs SNMT direct models}
The best multi-pivot result (27.1 BLEU) with an effective $\waitk$ of 16 is 4 BLEU points below the best direct translation result (31.1 BLEU) with a $\waitk$ of 8. Although this drop in translation quality is undesirable, consider the best single pivot results using French (21.3 BLEU) or Spanish (20.0 BLEU), which are 9.8 and 11.1 BLEU below the direct. Compared to the single pivot models, multi pivot models manage to recover most of the loss in translation quality. This shows that multiple pivots are more useful in an SNMT setting as compared to a full sentence setting. Once again, the reader should note that direct model scores are upper limits when parallel corpora exist, which may not always be the case.

\subsubsection{Comparing S2P And P2T Models}

When source and pivot models are both $\waitk$, the effective $\waitk$ is the sum of the individual $\waitk$’s. Consider an effective $\waitk$ of 8, which can be obtained by individual $\waitk$ combinations of (2, 6), (4, 4) and (6, 2) for source to pivot (S2P) and pivot to target (P2T). When French is used as the pivot, the corresponding scores are 16.1, 16.1 and 12.8 BLEU. In the case of Spanish, the scores are 14.7, 13.9 and 11.0. Finally, when both pivot languages are used, the scores are 18.7, 20.1 and 17.6. All these scores show that a longer $\waitk$ for the P2T NMT model is more important than a longer $\waitk$ for the S2P NMT model. This makes sense as the P2T model is responsible for the final translation and hence needs to have as much context as possible for translation. This might also mean that the P2T model is the bottleneck, and multi-pivot NMT addresses a part of this bottleneck. In the future, we will investigate how to reduce the need for long $\waitk$'s for P2T models.

\section{Conclusion}
This paper proposed multi-pivot NMT and extended it to a simultaneous NMT setting where the source to pivot and pivot to target models are $\waitk$ simultaneous NMT models. We showed that in a full sentence setting, multiple pivots can lead to an improvement of up to 2.4 BLEU and in an SNMT setting, multiple pivots can lead to an improvement of up to 5.8 BLEU. This shows that multiple pivots are even more critical in real-time translation settings. The work presented in this paper, due to its preliminary nature, relies on N-way corpora for multi-source (pivots in our work) translation, however previous work on multi-source translation \cite{firat16b} has shown that we do not need N-way corpora at all. Our future work will focus on obviating the need for N-way corpora as well as pushing the limits of translation quality via a larger number of pivot languages.
\bibliography{emnlp2021}
\bibliographystyle{acl_natbib}

\end{document}